\documentclass[10pt,twocolumn,letterpaper]{article}
\pdfoutput=1
\usepackage{cvpr}
\usepackage{times}
\usepackage{epsfig}
\usepackage{graphicx}
\usepackage{amsmath}
\usepackage{amssymb}
\usepackage{subfig}

\usepackage{color}

\usepackage[breaklinks=true,bookmarks=false]{hyperref}

\cvprfinalcopy 

\def\cvprPaperID{****} 

\DeclareMathOperator*{\argmax}{arg\,max}

\begin{document}

\title{Event Specific Multimodal Pattern Mining with Image-Caption Pairs}

\author{Hongzhi Li\thanks{Co-First Author. Hongzhi Li and Joseph Ellis contributed equally to this paper.}\\
Columbia Univeristy\\
{\tt\small hongzhi.li@columbia.edu}
\and
Joseph G. Ellis\footnotemark[1]\\
Columbia Univeristy \\
{\tt\small jge2105@columbia.edu}
\and
Shih-Fu Chang\\
Columbia Univeristy\\
{\tt\small sfchang@ee.columbia.edu}
}
\maketitle

\begin{abstract}
In this paper we describe a novel framework and algorithms for discovering image patch patterns from a large corpus of weakly supervised image-caption pairs generated from news events.
Current pattern mining techniques attempt to find patterns that are representative and discriminative, we stipulate that our discovered patterns must also be recognizable by humans and preferably with meaningful names.
We propose a new multimodal pattern mining approach that leverages the descriptive captions often accompanying news images to learn {\it semantically meaningful} image patch patterns.
The mutltimodal patterns are then named using words mined from the associated image captions for each pattern.
A novel evaluation framework is provided that demonstrates our patterns are 26.2\% more semantically meaningful than those discovered by the state of the art vision only pipeline, and that we can provide tags for the discovered images patches with 54.5\% accuracy with no direct supervision.
Our methods also discover named patterns beyond those covered by the existing image datasets like ImageNet.
To the best of our knowledge this is the first algorithm developed to automatically mine image patch patterns that have strong semantic meaning specific to high-level news events, and then evaluate these patterns based on that criteria.
\end{abstract}

\begin{figure}
  \centering
  \includegraphics[width=0.45\textwidth]{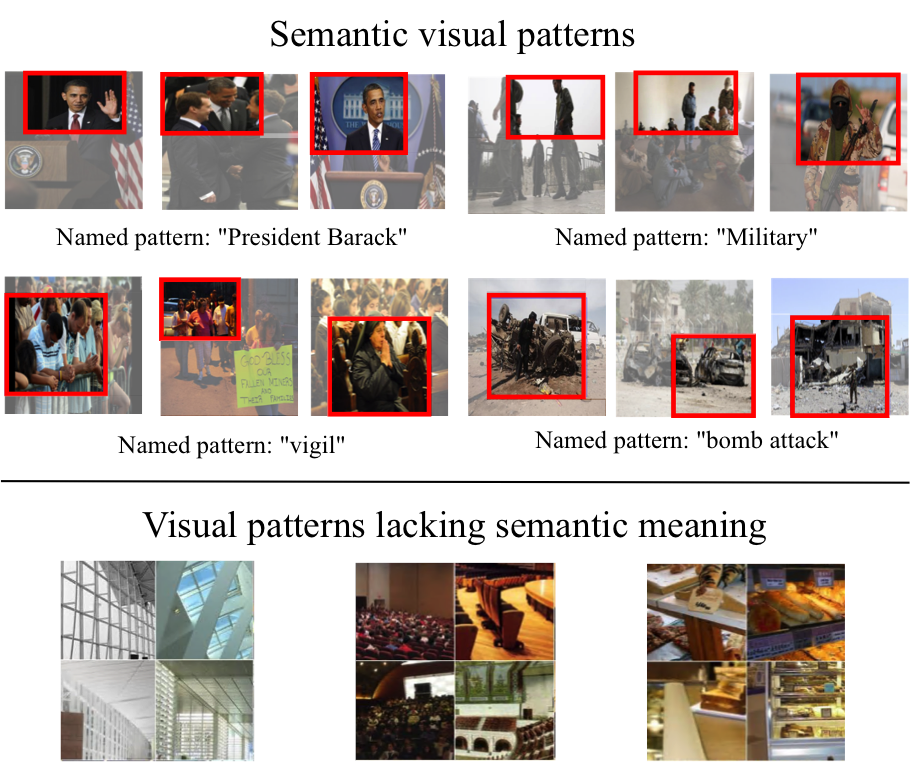}
  \caption{Examples of visual patterns generated from different approaches. The top are our multimodally mined visual patterns with names generated from our news event dataset, and the bottom are generated using the current state-of-the-art visual pattern mining approach on the MITIndoor dataset \cite{LiLSH15CVPR}.}
  \label{fig:patterns}
\end{figure}

\section{Introduction}
With recent advances in the capabilities of computer vision systems, researchers have been able to demonstrate performance at near-human level capabilities in difficult tasks such as image recognition.
These advances have been made possible by the pervasiveness of deep learning network architectures that utilize vast amounts of supervised image information.
Image classification and recognition performance on supervised datasets will continue to improve, and because of this rapid improvement technologies are quickly being integrated into production level products such as Google Photos with tremendous public approval.
Now that we can perform supervised image classification and recognition tasks with great accuracy we propose to utilize the developed technologies to attempt tasks that may not be as straightforward.
We propose one such problem: how can we discover semantically meaningful image patches from a large set of unstructured image and caption pairs? In addition, how do we discover meaningful patterns specific to high-level events such as ``Baltimore Protest''.

Recently, \cite{singh2012unsupervised} and \cite{LiLSH15CVPR} proposed using image patches as a feature for classification instead of low-level visual bag of words features (BoW).
Visual patterns have also been used to summarize image collections by \cite{zhang2014scalable}.
Traditionally, the desired visual patterns must be {\it representative} and {\it discriminative} to perform well on image classification or summarization.
{\it Representative} refers to the tendency of this patch to appear in many of the images for a target class, and {\it discriminative} refers to the fact that it should be unique and not appear frequently in images outside of the target category.
Insuring that the patterns found are representative and discriminative leads to patterns that are useful for image classification, but does not imply that they are meaningful on their own or useful for other high level information extraction tasks.
Unique but often meaningless local texture-based image patches are usually discovered using traditional pattern mining methods.
If image patterns are semantically meaningful they can be used for tasks beyond just image classification.
For example, the image patches can be used to automatically understand the possible actors and actions in a collection of images from a specific scenario or event.
This information can actually be mined from the data, eliminating the need for expensive supervised bounding box based training data for learning.
Our goal in this paper is to discover image patches that are not only discriminative and representative, but also contain high-level semantic meaning and are not simply a particular texture or part of a larger object.
We establish a third important characteristic of our image patches, that our discovered patches must be {\it informative}.
We define a {\it semantic visual pattern} as the patterns that are recognizable to a human and can be explained or named by a word or collection of words.
Fig.~\ref{fig:patterns} gives an example of the difference between our generated {\it semantic visual patterns} and the vision only approach.

To accomplish this goal we have created an image-caption pair dataset with categories that are news related events such as a {\it demonstration} or {\it election}, which can be visually diverse.
This visual diversity of news events allows us to discover higher level concepts with our image patches compared to previous pattern mining works which operate on datasets for scene classification or image recognition with lower level categories.
For example, in our dataset we are able to find image patches that represent concepts like air strike, press conference, or award ceremony.
Contrast this with the state of the art on the MIT Indoor scene recognition dataset \cite{LiLSH15CVPR}, where the discovered patterns correspond to a chair, ceiling or other very concrete lower level concepts.
To generate image patch patterns that are informative we propose a new multimodal pattern mining algorithm, which will be shown to generate patterns that are more semantically meaningful than visual-only methods in Sec.~\ref{sec:sematic_eval}.
Due to the fact that we utilize image captions we are also able to name our discovered image patch patterns.
To the best of our knowledge, no other related works address the issue or evaluate whether their discovered image patch patterns are informative.
Other researchers use the discovered patterns to build a mid-level representation of images and test the new features on image classification tasks.
These evaluations demonstrate that the mid-level visual patterns are certainly useful, but in our opinion these evaluations don't address the fundamental question of pattern mining, what is the pattern?
We argue that the current evaluation procedures are not sufficient to evaluate the quality of discovered patterns.
In this paper, we propose a set of new subjective methods in conjunction with the objective methods to evaluate the performance of visual pattern mining.
The contributions of our work are as follow:
\begin{itemize}
\item To the best of our knowledge, we are the first to address the creation of high-level event specific image patch patterns that are semantically meaningful.
\item Demonstrate a novel multimodal pattern mining approach using image-caption pairs and show that using caption data we are able to discover image patches that are recognizable and apply meaningful names to them.
\item Show that directly using the output of the last pooling layer of a CNN network we are able to obtain a $25\times$ reduction in computational cost than when using the state of the art image patch sampling approach provided in \cite{LiLSH15CVPR} with little performance drop off in visual transaction generation for pattern mining.
\item Provide an evaluation framework for determining if our discovered patterns are semantically informative.
Our performance demonstrates that we can learn semantically meaningful patterns from large weakly supervised datasets and name them using our methods.
\item We will make our collected image-caption dataset and named image patch patterns available to the public at the time of publication. We will also release the tools for mining the namable patterns so that other researchers will be able to extend the methods to other data corpora.
\end{itemize}

\section{Related Work}
Low level image features such as SIFT \cite{Lowe:2004:DIF:993451.996342} and Bag-of-word methods were widely used as a representation for image retrieval and classification.
However, researchers have proven that these low level features do not have enough power to represent the semantic meaning of images.
Mid-level image feature representations are often used to achieve better performance in a variety of computer vision tasks.
Some frameworks for using middle level feature representation, such as \cite{li2010object}, \cite{TorresaniSzummerFitzgibbon10}, have achieved excellent performance in object recognition and scene classification.
ImageNet \cite{deng2009imagenet}, was introduced and has lead to breakthroughs in tasks such as object recognition and image classification due to the scale of well-labeled and annotated data.
Each of the images within Image-Net are manually labeled and annotated, this is a very expensive and time-consuming task.
This work looks beyond a manually defined ontology, and instead focuses on mining mid-level image patches automatically from very weakly supervised data to attempt to unbound researchers from the need for costly supervised datasets.
We approach this problem from a multi-modal perspective (using the image and caption together), which allows us to name and discover higher level image concepts.

Visual pattern mining is an important task since it is the foundation of many middle-level feature representation frameworks.
\cite{han2000mining} and \cite{zhang2014scalable} use low level features and a hashing approach to mine visual patterns from image collections.
\cite{yuan2007spatial} utilizes a spatial random partition to develop a fast image matching approach to discover visual patterns.
All of these methods obtain image patches from the original image collection either by random sampling or salient/objectness detection and utilize image matching or clustering to detect similar patches to create visual patterns.
These methods are computationally intense, because they have to examine possibly hundreds or thousands of image patches from each image.
These methods rely heavily on low level image features, and therefore do not often produce image patches that exhibit high level semantic meaning.
The generated image patterns are usual visually duplicated or near-duplicated image patches.

Convolutional neural networks (CNN) have achieved great success in many research areas \cite{Simonyan14c}, \cite{Alexnet}.
Recently, \cite{LiLSH15CVPR} combined the image representation from a CNN and association rule mining technology to effectively mine visual patterns.
They first randomly sampled image patches from the original image and extracted the fully connected layer response as features for each image patch utilized in an association rule mining framework.
This approach is able to find consistent visual patterns, but cannot guarantee the discovered visual patterns are semantically meaningful.
Most of the existing visual pattern mining work focuses on how to find visually consistent patterns.
To the best of our knowledge, our paper is the first attempt to find high level semantically meaningful visual patterns.

Another category of related works is image captioning.
In recent years, many researchers have focused on teaching machines to understand images and captions jointly.
Image caption generation focuses on automatically generating a caption that directly describes the content in an image using a language model.
Multimodal CNNs \cite{frome2013devise} or RNN \cite{mao2014explain} frameworks are often used to generate sentences for the images.
All the existing work use supervised approaches to learn a language generation model based on carefully constructed image captions created for this task.
The datasets used in caption generation, such as the MSCoco dataset \cite{MSCoco} consist of much simpler sentences than appear in news image-caption pairs.
We differ from these approaches in that we do not try to generate a caption for images, but instead use them jointly to mine and name the patterns that appear throughout the images.

\section{Multimodal Pattern Mining}
In this section we discuss our multimodal pattern mining approach.
In particular, how we collected a large scale dataset, generated feature based transactions from the image and captions, and how we find semantic visual patterns and name them.

\begin{table}
\centering
\caption{Number of images per event category for some of the most popular event categories in our dataset.}
\begin{tabular}{ll|ll}
\hline
\hline
Event & \# of Images & Event & \# of Image \\ \hline
Attack & 52649 & Injure & 5853 \\
Demonstrate & 20933 & Transport & 51187 \\
Elect & 9265 & Convict & 1473 \\
Die & 26475 & Meet & 32787 \\ \hline
\end{tabular}
\label{tab:event_nums}
\end{table}

\subsection{Weakly Supervised Event Dataset Collection}
We believe that by using weakly supervised image data from target categories that are sufficiently broad we can automatically discover meaningful and recongizable image patch patterns.
To accomplish this task we collect a set of image caption pairs from a variety of types of news event categories.
Fig.~\ref{fig:patterns} provides an example of the differences and variability of the visual content between news images and a scene classification dataset that allow us to learn higher level image patterns.

We begin by crawling the complete Twitter feeds of four prominent news agencies, the Associated Press, Al Jazeera, Reuters, and CNN.
Each of these agencies have a prominent twitter presence, and tweet links to their articles multiple times a day.
We collect the links to the articles and then download the html file from each extracted link.
The article links span the time frame from 2007-2015, and cover a variety of different topics.
We then parse the raw html file and find the image and caption pairs from the downloaded news articles.
Through this process we are able to collect approximately 280k image-caption pairs.

Once we have collected the dataset we want to find image-caption pairs that are related to different events covered in news.
We utilized the event ontology that was defined for the Knowledge Base Population (KBP) task in the National Institute for Standards and Technology Text Analysis Conference in 2014 to provide supervision to our dataset.
Within this task there is an event track with the stated goal to ``extract information such that the information would be suitable as input to a knowledge base.''
This track goal closely models the goals of learning patterns that are recognizable and hence could be used in knowledge base population.
Making this ontology a perfect fit for our task.

The KBP event task utilizes the ontology defined by the Lingustic Data Consortium in 2005 \cite{ACE}.
This event ontology contains 34 distinct event types, the events are broad actions that appear commonly throughout news documents, such as {\it demonstrate}, {{\it divorce}, and {\it convict}.
Provided in the training data with these event ontologies is a list of trigger words for each of the events that are used to detect when an event appears in text data.
An example of some of the trigger words used for the {\it demonstrate} event are: protest, riot, insurrection, and rally.
We search each of the captions for a trigger word from the event category, and if an image caption contains that trigger word we assign that image caption pair to the given event category.
The number of images found for some of the event categories can be seen in Table.~\ref{tab:event_nums}.

\subsubsection{Review of Pattern Mining}
\label{pattern_review}
In this section we will review the basic ideas and definitions necessary for pattern mining.
Assume that we are given a set of $n$ possible observations $X = \{ x_1, x_2, ... x_n \}$, a {\it transaction}, $T$, is a set of observations such that $T \subseteq X$.
Given a set of transactions $S = \{T_1, T_2, ... T_m\}$ containing $m$ transactions, our goal is to find a particular subset of $X$, say $t^*$, which can accurately predict the presence of some target element $y \in T_a$, given that $t^* \subset T_a$ and $y \cap t^* = \emptyset$.
$t^*$ is referred to as a {\it frequent itemset} in the pattern mining literature.
The relationship from $t^* \rightarrow y$ is known as an {\it association rule}.
The support of $t^*$ reflects how often $t^*$ appears in $S$ and is defined as,
\begin{equation}
s(t^*) = \frac{|\{T_{a} |t^* \subseteq T_{a},  T_{a} \in S \}|}{m}
\label{eq:st}
\end{equation}

Our goal is to find association rules that accurately predict the correct event category for the image-caption pairs.
Therefore, we want to find patterns such that if $t^*$ appears in a transaction there is a high likelihood that $y$, which represents an event category, appears in that transaction as well.
We define the {\it confidence} as the likelihood that if $t* \subseteq T$ then $y \in T$, or,
\begin{equation}
\label{confidence}
c(t^* \rightarrow y) = \frac{s(t^* \cap y)}{s(t^*)}.
\end{equation}

\subsubsection{Using CNN Pool5 Features for Transaction Generation from Images}
\label{pool5}
Certain portions of a CNN are only activated by a smaller region of interest (ROI) within the original image.
The last layer in which the particular neurons do not correspond to the entire image is the output of the final convolutional and pooling layer for \cite{Alexnet}.
Based on this observation, for each image we find the maximum magnitude response from a particular feature map from the pool5 layer of the CNN defined by \cite{Alexnet}.
The pool5 layer of this network consists of 256 filters, and the response of each of the filters over a $6 \times 6$ mapping of the image.
The corresponding ROI from the original image in which all the pixels in that region contribute to the response of a particular neuron in the pool5 layer is a $196 \times 196$ image patch from the $227 \times 227$ resized image.
These $196 \times 196$ image patches come from a zero-padded representation of the image with zero-padding around the image edges of $64$ pixels and a stride of $32$.
Namely, from a $227 \times 227$ scaled input image, a total of $6 \times $6 (36) patch areas are covered from all the stride positions, resulting in a $6 \times 6$ feature map for each filter in the layer 5.
Using this approach we are able to use the current existing architecture to compute the filter responses for all patches at once without actually changing the network structure.
This idea allows us to extract image patch patterns in a way that is much more efficient than current state of the art methods.

We use the pre-trained CNN model from \cite{Alexnet} that was trained using the ImageNet dataset to extract the pool5 features for the news event images.
For each image, we keep the maximum response over the $6 \times 6$ feature map and set other non-maximal values to zero for all $256$ filters, similar to the non-maximum suppression operation in the literature. Such an operation is used to find the patch triggering the highest response in each filter and avoid redundant patches in the surrounding neighborhood in the image that may also generate high responses. The above process results in a $256$ dimensional feature vector representation for each image patch.
It has been shown in \cite{LiLSH15CVPR} that most of the information is carried by the top $k$ magnitude dimensions of the feature vector, and the magnitude response of these dimensions is not particularly important.
Thus, we follow their suggestion and set the top $k$ magnitude dimensions of the CNN vector to 1 and the others to 0, creating a binary representation of which filters are activated for each image patch.
We use these sparse binarized features to build {\it transactions} for each image patch as discussed in Sec.~\ref{pattern_review}, where the non-zero dimensions are the items in our transaction.

\begin{figure}
  \centering
  \includegraphics[width=0.45\textwidth]{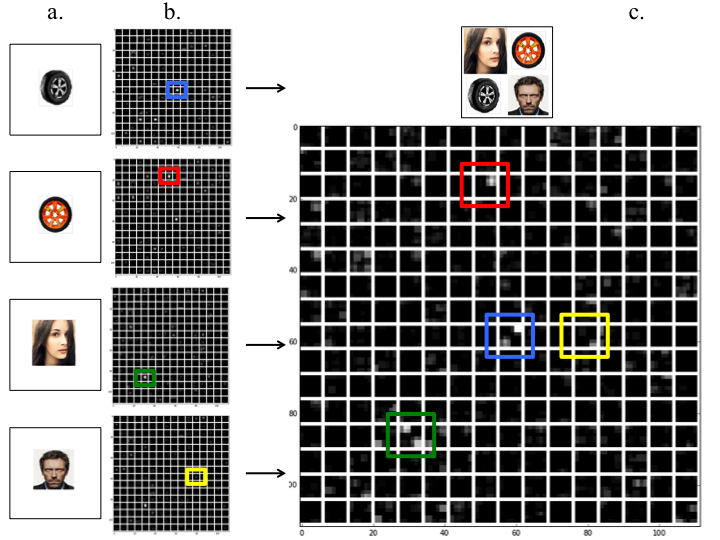}
  \caption{Localization power of pool5.  In column a.) are example image patches placed at the middle of an image, column b.) represents the pool5 response map of all 256 filters over the spatial resolution, and column c.) shows an example where all 4 image patches are placed in an image together and its respective pool5 feature response map.
  }
  \label{fig:pool5}
\end{figure}

By utilizing the architecture of the CNN directly we are able to efficiently extract image features that come from specific ROI that are suitable for pattern mining.
The current state of the art pattern mining technique proposed by \cite{LiLSH15CVPR} requires a sampling of the image patches within each image and then operating the entire CNN over this sampled and resized image patch.
This procedure is very costly, because the CNN must be used to extract features from a number of sampled images that can be orders of magnitude larger than the dataset size.
For example, for the MIT indoor dataset the authors of \cite{LiLSH15CVPR} sample $128 \times 128$ size image patches with a stride of 32 from images that have been resized such that their smallest dimension is of size 256.
Thus, the number of image samples that are taken for each image is greater than or equal to $(\frac{256 - 128}{32} + 1)^2 = 25$.
The full CNN must operate on all of these sampled images.
In contrast our method works directly on the image themselves without any sampling.
We are able to extract representations for 36 image patches from an image while only having the CNN operate on the image once.
By leveraging the structure of the CNN during test or deployment our method is {\it at least} $25$ times less computationally intensive than the current state of the art.
We demonstrate in Sec.~\ref{classification_eval} that we are able to achieve this computational gain with very little cost to predictive performance.
An example of the localization power of pool5 features for pattern mining can be seen in Fig.~\ref{fig:pool5}.

\subsubsection{Generating Transactions from Captions}
We have discussed how we generated transactions by binarizing and thresholding the CNN features that are extracted from the images.
We need an analogous algorithm for generating transactions from image captions.

We begin by cleaning each of the image captions by removing stopwords and other ancillary words that should not appear (html tags or urls).
We then tokenize each of the captions and find all of the words that appear in at least 10 captions in our dataset.
Once we find these words we use the skip-gram model proposed in \cite{DBLP:journals/corr/MikolovSCCD13} that was trained on a corpus of Google News articles to map each word to a 300 dimensional embedded space.
The skip-gram model works well in our context because words with similar uses end up being embedded close to each other in the feature space.
Words such as ``religious clergy'', ``priest'', and ``pastor'' all end up close in euclidean distance after embedding and far away from words that are not similar.
We cluster the words using K-means clustering to generate 1000 word clusters.

To generate transactions for each caption we map each word back to its corresponding cluster, then include that cluster index in the transaction set.
We remove patterns that contain cluster indices that are associated with commonly used words by having a high confidence score threshold as defined in Eq.~\ref{confidence}.
The cluster indices that frequently appear in captions from a particular event category but rarely for other categories are found through our association rule mining framework.

We require our discovered patterns to contain items from both the visual and text transactions.
By requiring words with similar meaning to appear in all captions of a visual pattern we are able to discard patterns that may be visually similar but semantically incoherent.
The skip-gram based algorithm is able to handle differences in word choice and structure between captions to effectively encode meaning into our multimodal transactions.

\subsection{Mining the Patterns}
We add the event category of each image as an additional item in the generated transaction for each of the image caption pairs.
Inspired by \cite{LiLSH15CVPR}, we use the popular apriori algorithm \cite{Agrawal:1994:FAM:645920.672836} to find patterns within the transactions that predict which event category the image belongs to.
We only find the association rules which have a confidence higher than $0.8$, and calculate the support threshold that insure that the at least $30$ image patches exhibit a found pattern.
Finally, we also remove any rules which only contain items generated from the image or caption transactions, insuring that we only retain truly multimodal patterns.
Therefore, our pattern requirements can be described mathematically as,

\begin{align}
&c(t^* \rightarrow y) \geq c_{min} \nonumber \\
&s(t^*) \geq s_{min} \nonumber \\
&t^* \cap \mathbf{I} \neq \emptyset \nonumber \\
&t^* \cap \mathbf{C} \neq \emptyset.
\label{eq:contstraints}
\end{align}

Where as defined in Eq \ref{eq:st} and Eq \ref{confidence}, y is the event category of the image-caption pair, $c_{min}$ is the minimum confidence threshold, $s_{min}$ is our minimum support threshold, $\mathbf{I}$ represents the items generated from the image transaction pipeline, and $\mathbf{C}$ are those generated from the caption pipeline. At the end, each multimodal pattern t* contains a set of visual items (fired filter responses in pool5 in CNN model) and a set of text patterns (clusters in the skip-gram embedded space).

\subsection{Naming the patterns}
If we can name the generated patterns we can use them for higher level information extraction tasks.
We leverage the fact that we have captions associated with each of the images to generate names for each pattern.

We begin the process of name generation by removing the words that are not generally useful for naming but appear often in captions.
The words that are removed include standard English language stop words (or, I, etc.), the name of each month, day, and directional words such as ``left'' and ``right''.
After cleaning the caption words we then encode both unigram and bigrams into a vector using tf-idf encoding.
We ignore any unigram or bigram that does not appear at least 10 times across our caption dataset.

Once these words are removed we then sum the tf-idf vector representations of each word in all of the captions associated with a particular pattern.
We then take the argument max over the summed tf-idf representations to obtain the name for this pattern.
This procedure is explained mathematically in the following way:
Let $p$ be a found multimodal itemset (pattern), and $T_k$ is the multimodal transaction for the $k$'th generated transaction in our dataset.
We define the set $P$ as all the indices of transactions that contain $p$, or $P = \{i|p \subseteq T_i, \forall i\}$.
In Eq.~\ref{namingtf-idf}, $\mathbf{V}$ is our vocabulary, $W_k$ is the set of words from the $k$'th caption, $\mathbf{I}_p(w)$ is an indicator function on whether $w$ corresponds to a cluster in the itemset of $p$, and $w_{kj}$ is the $j$'th word in the $k$'th caption,

\begin{equation}
w_{name} = \argmax_{w \in \mathbf{V}} \sum_{k \in P} \sum_{w_{kj} \in W_k} \mathbf{I}_p(w_{kj})*tfidf(w_{kj})
\label{namingtf-idf}
\end{equation}

\begin{figure}[t]
  \centering
  \includegraphics[width=0.5\textwidth]{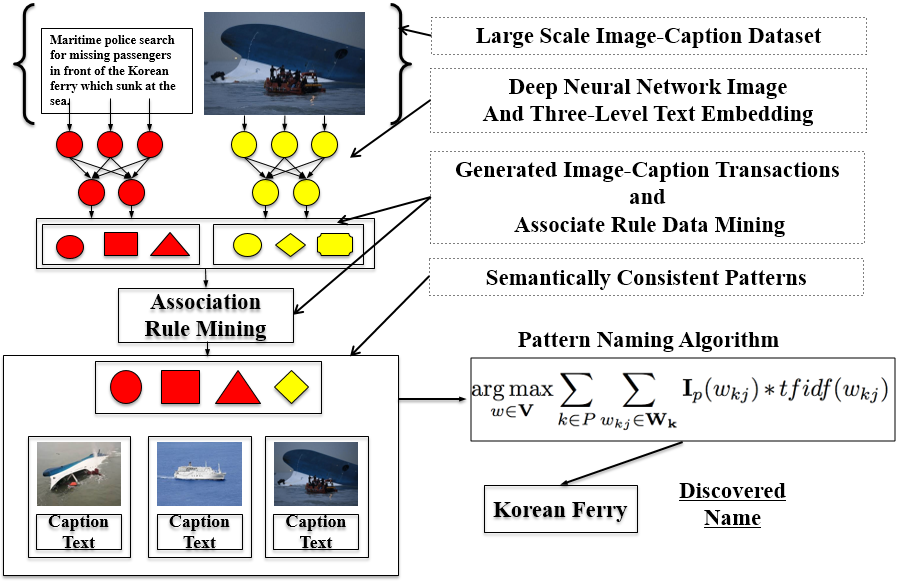}
  \caption{Our full pattern discovery and naming pipeline}
  \label{fig:naming_pipeline}
\end{figure}

Once the names are found, we remove any name that appears in more than 10\% of the captions of a particular event.
This is important because for particular events like ``injure'', words such as {\it injure} and {\it wounded} appear across many captions, and may lead to poor naming.
Some examples of discovered patterns and the names that we have assigned to them can be seen in Fig~\ref{fig:naming}.
Our full pattern naming algorithm and pipeline can be seen in Fig~\ref{fig:naming_pipeline}.

\section{Evaluation}
\subsection{Using visual patterns for classification}
\label{classification_eval}
The most popular evaluation for pattern mining is to produce a middle level feature representation for the images, and then use any supervised classification approach to perform a task.
The authors of \cite{LiLSH15CVPR} take this approach.
Instead of producing a mid-level visual feature, we integrate our discovered patterns into a CNN and perform the image classification task using a modified CNN network.

We modify a CNN \cite{Alexnet} by adding a fixed fully connected layer on top of the last max pooling layer, pool5.
The max pooling layer produces an $256$ ($N$) dimensional output for each image patch.
The fixed fully connected layer has $M*N$ parameters, where $M$ is the number of patterns we mined in the training dataset using the visual transactions and the apriori algorithm.
Each dimension in the $M$-dimensional parameter vector corresponds to one of our mined patterns.
The fixed fully connected layer for our model is a sparse matrix with 1 on the dimension of the filters present in a pattern, and 0 elsewhere.
This fixed fully connected layer produces an $M$ dimensional output which corresponds to whether a pattern is present in the image patch or not.
On top of this pattern detection layer, we add another fully connected layer, followed by a ReLU layer and softmax layer to perform image classification.
We share the same convolutional layers as \cite{Alexnet}, therefore we use a pre-trained model to update the parameters of all the convolution layers.
During the training process, we freeze all the parameters of convolution layers and the first fully connected layer.
Only the second fully connected layer and the softmax layer are modified during training.
We compare our results with the current state-of-the-art methods in table \ref{tab:MITIndoor} on the commonly used MIT Indoor dataset \cite{10.1109/CVPRW.2009.5206537}.
This dataset is widely used to evaluate performance in pattern mining tasks.
We can see that our model outperforms all but \cite{LiLSH15CVPR}.
However, our model is at least $25$ times more efficient in transaction generation than their model with comparable performance.

\begin{table}
\centering
\caption{Scene classification results on MITIndoor dataset}
\begin{tabular}{lll}
\hline
\hline
Method                       & Accuracy (\%)           \\ \hline
ObjectBank \cite{li2010object}                                  & 37.60                  \\
Discriminative Patch \cite{singh2012unsupervised}                     & 38.10                  \\
BoP \cite{juneja2013blocks}                                & 46.10                  \\
HMVC \cite{li2013harvesting}                                  & 52.30                 \\
Discriminative Part \cite{sun2013learning}                 & 51.40                  \\
MVED \cite{doersch2013mid}                  & 66.87                  \\
MDPM \cite{LiLSH15CVPR}                                     & 69.69                  \\
ours                                      &  68.22        \\ \hline
\end{tabular}
\label{tab:MITIndoor}
\end{table}

\subsection{Multimodal Pattern Discovery}
Given our large news event image-caption dataset it is interesting to analyze how many multimodal patterns that we can find per event.
The number of patterns that were found for each event can be seen in Table~\ref{tab:pattern_nums}.
We can see that some events that are sufficiently visually and semantically varied will have many patterns discovered.
Other events that contain images that are all very similar such as {\it convict}, which oftentimes take place in a court room and generally only show people's faces have little or no found patterns.

We notice that although events with more images associated with them generally have more discovered patterns, this is not a concrete relationship.
For example, the {\it meet} event has almost 10 times as many discovered patterns as the {\it attack} category even though {\it attack} has more images associated with it.
When analyzing the data we see that many different summits, press conferences, and other types of meetings that are covered within the news media exhibit strong visual consistency.
Many images of word leaders shaking hands, talking behind podiums, and similar scenarios exist that we are able to mine as patterns that define this event.
Contrast this with the {\it attack} event which contains highly varied visual action content and we are able to find less visually consistent patterns.

This makes sense intuitively, because an attack can happen in a variety of ways, such as an air strike, shooting, bomb, riot, fire, etc.
The {\it Meet} event, on the other hand, almost always entails images of two or more people simply standing around and talking, and the background, setting, and people are often consistent across multiple images from one real scenario.
This allows us to create patterns from a variety of images from the same scenario, which exhibit strong visual and semantic consistency.
In this way we are able to find both very broad patterns in events such as {\it attack}, or more specific one time only patterns to describe things like the {\it republican convention} or {\it world health summit} in the {\it meet} event.
Both types of patterns are interesting and useful for information extraction.

We note that our methodology is useful in a variety of scenarios where weak labels can be easily obtained.
For example, our multimodal algorithm could also be applied to images from movies and their transcripts with genre level weak supervision.

\begin{table}
\centering
\caption{Number of found patterns in news events.}
\begin{tabular}{ll|ll}
\hline
\hline
Event & \# of Patterns & Event & \# of Patterns \\ \hline
Attack & 573 & Convict & 0 \\
Demonstrate & 1247 & Die & 146 \\
Elect & 42 & Injure & 45 \\
Meet & 5159 & Transport & 509 \\ \hline
\end{tabular}
\label{tab:pattern_nums}
\end{table}

\begin{figure*}[t]
\captionsetup[subfigure]{labelformat=empty}
\centering
\begin{tabular}{ccc | ccc | ccc}
\subfloat{\includegraphics[width = 0.5in,height=0.5in]{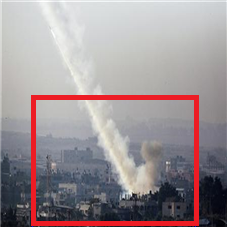}} &
\subfloat[Air Strike]{\includegraphics[width = 0.5in,height=0.5in]{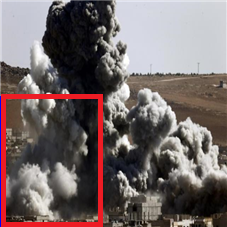}} &
\subfloat{\includegraphics[width = 0.5in,height=0.5in]{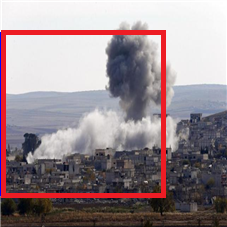}} &
\subfloat{\includegraphics[width = 0.5in,height=0.5in]{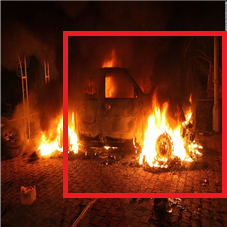}} &
\subfloat[Burn]{\includegraphics[width = 0.5in,height=0.5in]{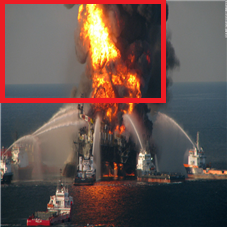}} &
\subfloat{\includegraphics[width = 0.5in,height=0.5in]{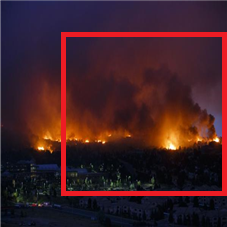}} &
\subfloat{\includegraphics[width = 0.5in,height=0.5in]{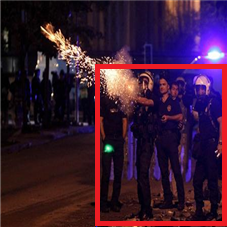}} &
\subfloat[Riot Police]{\includegraphics[width = 0.5in,height=0.5in]{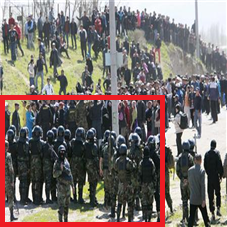}} &
\subfloat{\includegraphics[width = 0.5in,height=0.5in]{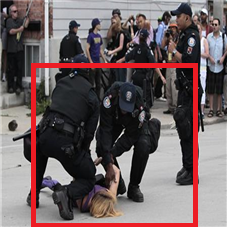}} \\ \hline
\subfloat{\includegraphics[width = 0.5in,height=0.5in]{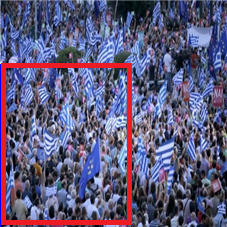}} &
\subfloat[Flags]{\includegraphics[width = 0.5in,height=0.5in]{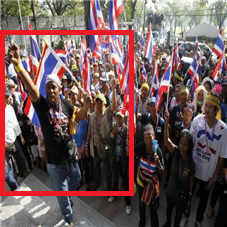}} &
\subfloat{\includegraphics[width = 0.5in,height=0.5in]{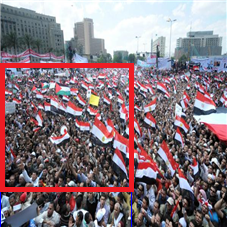}} &
\subfloat{\includegraphics[width = 0.5in,height=0.5in]{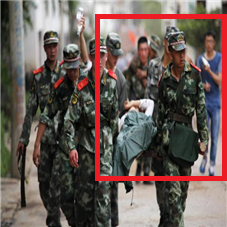}} &
\subfloat[Stretcher]{\includegraphics[width = 0.5in,height=0.5in]{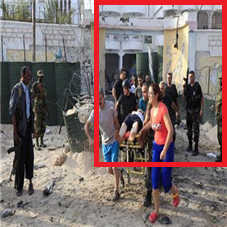}} &
\subfloat{\includegraphics[width = 0.5in,height=0.5in]{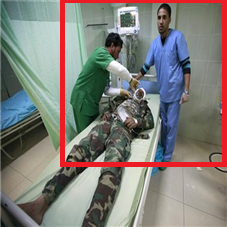}} &
\subfloat{\includegraphics[width = 0.5in,height=0.5in]{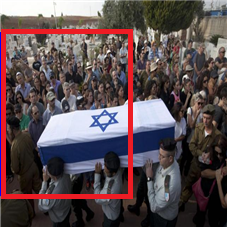}} &
\subfloat[Coffin]{\includegraphics[width = 0.5in,height=0.5in]{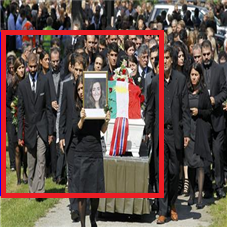}} &
\subfloat{\includegraphics[width = 0.5in,height=0.5in]{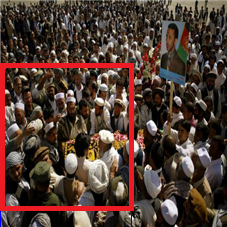}}
\end{tabular}
\caption{Examples of image patches and names discovered by our multimodal approach. The air strike and burn patterns were generated from the {\it attack} event, riot police and flags from {\it demonstration}, and stretcher and coffin from {\it injure}.}
\label{fig:naming}
\end{figure*}

\subsection{Subjective Analysis of Multimodal Patterns}
The type of evaluation in \ref{classification_eval} is useful in determining if the patterns are discriminative and representative, but this does not actually determine if the patterns are informative.
We believe that the pattern mining algorithm that we have developed can be used for more than image classification performance.
To prove this we have developed an evaluation to determine the percentage of our patterns that are semantically meaningful as well as if we are able to accurately name the patterns that we have found.

\subsubsection{Semantic Relevance of Patterns}
\label{sec:sematic_eval}
To accomplish this task we built an Amazon Mechanical Turk (AMT) interface, which we used to gather annotations.
To generate the patterns for inspection we randomly sampled patterns from the event categories of {\it Attack}, {\it Demonstrate}, {\it Transport}, {\it Injure}, and {\it Elect}, and {\it Meet} for a total of $99$ randomly sampled patterns.
These samples were taken from our multimodal pattern mining algorithm and a visual-only version of the algorithm, which is the same as the multimodal algorithm without using any text transactions.
We then randomly sample up to 16 of the image patches from a given pattern to represent the pattern and show them to the annotator.
The annotator is asked to decide whether the group of images exhibit similar and consistent semantic content.
If the annotator agrees that the images exhibit consistent semantics then we ask them to supply up to 3 tags that could be applied to this group of images.
We choose to have the annotators provide three tags (which can be synonyms), so that we have three words from each annotator to compare to each other given that different annotators may choose different tags with similar semantic meaning.
We analyze the inter-annotator tag agreement to determine if the patterns are semantically relevant.
Our assumption is that if the image patches exhibit strong semantic consistency then the annotators will supply the same tags for the group of images.
We obtained at least 4 annotations per pattern.

\begin{table}
\centering
\caption{Subjective Evaluation of semantic consistency.}
\begin{tabular}{ll}
\hline
\hline
Method & Accuracy \\ \hline
\% of Semantically Consistent Patterns \\
Multimodal & {\bf 82.0\%} \\
visual-only & 72.6\% \\ \hline
\% of Tag Agreement Among Subjects\\
Multimodal & {\bf 49.5\%} \\
visual-only & 39.2\% \\ \hline
\end{tabular}
\label{tab:semantic_meaning}
\end{table}

We can see the results from this experiment in Table~\ref{tab:semantic_meaning}.
Our multimodal model discovers patterns that are labelled by the annotators to be more semantically consistent than the visual-only patterns.
The annotators answered that a pattern generated by our multimodal method was semantically consistent $82\%$ of the time compared to $72.6\%$ for visual-only.
We believe that because humans have a strong capability for reasoning they may ``find'' semantic meaning in a group of images that could be spurious.
For this reason we also checked the agreement between the annotator provided tags.
If a pattern received tags from at least 3 annotators (annotators could say the patterns were not semantically consistent and choose not to provide tags), and {\it at least} half of the annotators provided the same tag for a pattern we decided that this pattern had ``tag agreement''.
In this metric our multimodal approach greatly outperformed the visual-only approach with the percentage of patterns exhibiting tag agreement for multimodal at {\bf 49.5\%} and visual-only at  39.2\%.
Our method outperforms the visual-only pipeline in all of our subjective tests, in particular we demonstrate that the agreement across a group of people of our image patches increases by {\bf 26.2\%} over the state of the art visual model.
The multimodal pipeline outperforms the visual-only model because the visual transaction pipeline can discover similar ``looking'' image patches across our image dataset, but these group of patches that ``look'' similar do not necessarily represent a similar semantic concept.
However, by adding our caption transaction generation to this pipeline we are able to discover image patches with consistent semantics.
For example, in the visual model pipeline we have patches taken from images of forest fires and explosions, because each pattern exhibits the similar visual components of smoke.
Our multimodal pipeline separates these images in a meaningful way, because a forest fire and explosion are semantically different concepts that may have little overlap in their text captions.
The multimodal information is particularly important in datasets like the one we have created because there is a high level of semantic and visual variance across the data.

\subsubsection{Evaluation of automatic pattern naming}
We test how well we are able to name the patterns.
To accomplish this task we utilize AMT to analyze whether our applied names are reasonable or not.
To generate the test data we utilized all of the patterns that were tested in Sec.~\ref{sec:sematic_eval}.
To evaluate our naming algorithm on an image patch level we asked an annotator to decide if the name that we supplied to the pattern described the semantic content.
From the 100 patterns we obtained 776 images.
We used 3 annotators per image-patch and name pair and took the majority vote to determine if that name aptly described the content in the image.
The annotation revealed that we had correctly applied names to 434 images and incorrectly to 353 images, for a total naming accuracy of {\bf 54.5\%}.
We remind the reader that this accuracy comes using {\it no} supervised data.
Our model is also able to learn patterns that are not contained in current supervised image datasets.
Using our naming algorithm we applied $246$ unique names to our patterns (some names apply to many patterns), but only $100$ ($40.6\%$) of these names are part of the current ImageNet ontology.
Examples of our mined semantic visual patterns and their applied names are shown in Fig.~\ref{fig:naming}.

\section{Conclusions}
We have developed the first dataset and algorithm for mining visual patterns from high level news event image caption pairs.
Our novel multimodal model is able to discover patterns that are more informative than the state of the art vision only approach, and accurately name the patterns.
This work represents the first approach in using multimodal pattern mining to learn high-level semantically meaningful image patterns.
The combination of our ability to find meaningful patterns and name them allow for many applications in high level information extraction tasks, such as knowledge base population using multimodal documents and automatic event ontology creation.


{\small
\bibliographystyle{ieee}
\bibliography{egbib}

\begin{thebibliography}{10}\itemsep=-1pt

\bibitem{Agrawal:1994:FAM:645920.672836}
R.~Agrawal and R.~Srikant.
\newblock Fast algorithms for mining association rules in large databases.
\newblock In {\em 20th International Conference on Very Large Data Bases},
  pages 487--499, 1994.

\bibitem{ACE}
e.~a. Christopher~Walker.
\newblock Ace 2005 multilingual training corpus.
\newblock In {\em Lingusitic Data Consortium}, 2006.

\bibitem{deng2009imagenet}
J.~Deng, W.~Dong, R.~Socher, L.-J. Li, K.~Li, and L.~Fei-Fei.
\newblock Imagenet: A large-scale hierarchical image database.
\newblock In {\em IEEE Conference on Computer Vision and Pattern Recognition
  (CVPR)}, pages 248--255, 2009.

\bibitem{doersch2013mid}
C.~Doersch, A.~Gupta, and A.~A. Efros.
\newblock Mid-level visual element discovery as discriminative mode seeking.
\newblock In {\em Advances in Neural Information Processing Systems (NIPS)},
  pages 494--502, 2013.

\bibitem{frome2013devise}
A.~Frome, G.~S. Corrado, J.~Shlens, S.~Bengio, J.~Dean, T.~Mikolov, et~al.
\newblock Devise: A deep visual-semantic embedding model.
\newblock In {\em Advances in Neural Information Processing Systems (NIPS)},
  pages 2121--2129, 2013.

\bibitem{han2000mining}
J.~Han, J.~Pei, and Y.~Yin.
\newblock Mining frequent patterns without candidate generation.
\newblock In {\em ACM SIGMOD Record}, pages 1--12, 2000.

\bibitem{juneja2013blocks}
M.~Juneja, A.~Vedaldi, C.~Jawahar, and A.~Zisserman.
\newblock Blocks that shout: Distinctive parts for scene classification.
\newblock In {\em IEEE Conference on Computer Vision and Pattern Recognition
  (CVPR)}, pages 923--930, 2013.

\bibitem{Alexnet}
A.~Krizhevsky, I.~Sutskever, and G.~E. Hinton.
\newblock Imagenet classification with deep convolutional neural networks.
\newblock In {\em Advances in Neural Information Processing Systems (NIPS)},
  pages 1097--1105. 2012.

\bibitem{li2010object}
L.-J. Li, H.~Su, L.~Fei-Fei, and E.~P. Xing.
\newblock Object bank: A high-level image representation for scene
  classification \& semantic feature sparsification.
\newblock In {\em Advances in Neural Information Processing Systems (NIPS)},
  pages 1378--1386, 2010.

\bibitem{li2013harvesting}
Q.~Li, J.~Wu, and Z.~Tu.
\newblock Harvesting mid-level visual concepts from large-scale internet
  images.
\newblock In {\em IEEE Conference on Computer Vision and Pattern Recognition
  (CVPR)}, pages 851--858, 2013.

\bibitem{LiLSH15CVPR}
Y.~Li, L.~Liu, C.~Shen, and A.~van~den Hengel.
\newblock Mid-level deep pattern mining.
\newblock {\em IEEE Conference on Computer Vision and Pattern Recognition
  (CVPR)}, pages 971--980, 2015.

\bibitem{MSCoco}
T.-Y. Lin, M.~Maire, S.~Belongie, J.~Hays, P.~Perona, D.~Ramanan,
  P.~Doll{\'a}r, and C.~L. Zitnick.
\newblock Microsoft coco: Common objects in context.
\newblock In {\em European Conference on Computer Vision (ECCV)}, 2014.

\bibitem{Lowe:2004:DIF:993451.996342}
D.~G. Lowe.
\newblock Distinctive image features from scale-invariant keypoints.
\newblock {\em Int. J. Comput. Vision}, 60(2):91--110, Nov. 2004.

\bibitem{mao2014explain}
J.~Mao, W.~Xu, Y.~Yang, J.~Wang, and A.~L. Yuille.
\newblock Explain images with multimodal recurrent neural networks.
\newblock {\em arXiv preprint arXiv:1410.1090}, 2014.

\bibitem{DBLP:journals/corr/MikolovSCCD13}
T.~Mikolov, I.~Sutskever, K.~Chen, G.~Corrado, and J.~Dean.
\newblock Distributed representations of words and phrases and their
  compositionality.
\newblock {\em Computing Research Repository (CoRR)}, abs/1310.4546, 2013.

\bibitem{10.1109/CVPRW.2009.5206537}
A.~Quattoni and A.~Torralba.
\newblock Recognizing indoor scenes.
\newblock {\em IEEE Conference on Computer Vision and Pattern Recognition
  (CVPR)}, pages 413--420, 2009.

\bibitem{Simonyan14c}
K.~Simonyan and A.~Zisserman.
\newblock Very deep convolutional networks for large-scale image recognition.
\newblock {\em Computing Research Repository (CoRR)}, abs/1409.1556, 2014.

\bibitem{singh2012unsupervised}
S.~Singh, A.~Gupta, and A.~Efros.
\newblock Unsupervised discovery of mid-level discriminative patches.
\newblock {\em European Conference on Computer Vision (ECCV)}, pages 73--86,
  2012.

\bibitem{sun2013learning}
J.~Sun and J.~Ponce.
\newblock Learning discriminative part detectors for image classification and
  cosegmentation.
\newblock In {\em IEEE International Conference on Computer Vision (ICCV)},
  pages 3400--3407, 2013.

\bibitem{TorresaniSzummerFitzgibbon10}
L.~Torresani, M.~Szummer, and A.~Fitzgibbon.
\newblock Efficient object category recognition using classemes.
\newblock In {\em European Conference on Computer Vision (ECCV)}, pages
  776--789, 2010.

\bibitem{yuan2007spatial}
J.~Yuan and Y.~Wu.
\newblock Spatial random partition for common visual pattern discovery.
\newblock In {\em IEEE International Conference on Computer Vision (ICCV)},
  pages 1--8, 2007.

\bibitem{zhang2014scalable}
W.~Zhang, H.~Li, C.-W. Ngo, and S.-F. Chang.
\newblock Scalable visual instance mining with threads of features.
\newblock In {\em ACM International Conference on Multimedia}, pages 297--306,
  2014.

\end{thebibliography}
}

\end{document}